# Default reasoning over domains and concept hierarchies

Pascal Hitzler

Department of Computer Science, Dresden University of Technology

**Abstract.** W.C. Rounds and G.-Q. Zhang have proposed to study a form of disjunctive logic programming generalized to algebraic domains [1]. This system allows reasoning with information which is hierarchically structured and forms a (suitable) domain. We extend this framework to include reasoning with *default negation*, giving rise to a new nonmonotonic reasoning framework on hierarchical knowledge which encompasses answer set programming with extended disjunctive logic programs. We also show that the hierarchically structured knowledge on which programming in this paradigm can be done, arises very naturally from formal concept analysis. Together, we obtain a default reasoning paradigm for conceptual knowledge which is in accordance with mainstream developments in nonmonotonic reasoning.

## 1 Introduction

In [1], Rounds and Zhang propose to study a form of clausal logic generalized to algebraic domains, in the sense of domain theory [2]. In essence, they propose to interpret finite sets of compact elements as clauses, and develop a theory which links corresponding logical notions to topological notions on the domain. Amongst other things, they establish a sound and complete resolution rule and a form of disjunctive logic programming over domains. A corresponding semantic operator turns out to be Scott-continuous.

We will utilize this proposal as a link between the formerly unrelated areas of formal concept analysis, on the one hand, and nonmonotonic reasoning, in the form of answer set programming, on the other. The relationships thus worked out serve a threefold purpose, namely (1) to obtain a sound domain-theoretic perspective on answer set programming, (2) to provide a formal link between domain logics and formal concept analysis for the purpose of cross-transfer of methods and results, and (3) to devise a reasoning paradigm which encompasses two formerly unrelated formalisms for commonsense reasoning, namely formal concept analysis, and answer set programming.

So in this paper, we will extend the logic programming paradigm due to Rounds and Zhang to include reasoning with default negation. We are motivated by the gain in expressiveness through the use of negation in artificial intelligence paradigms related to nonmonotonic reasoning. This approach, using ideas from default logic [3], treats negation under the intuition that the negation of an item shall be believed if there is no reason to believe the item itself. This perspective on negation has recently led to the development of applications in the form of nonmonotonic reasoning systems known as *answer set programming*, the two most popular probably being dlv and smodels [4,5]. We will indeed see that the extension of the approach by Rounds and Zhang by default negation is a natural generalization of answer set programming with extended disjunctive logic programs [6].

On the other hand, building on the work reported in [7], we establish a strong connection between the clausal logic on algebraic domains mentioned above, and fundamental notions from formal concept analysis [8]. More precisely, we will see that in certain cases the formation of formal concepts from formal contexts can be recast naturally via the notion of logical consequence in Rounds' and Zhang's clausal logic. Our default reasoning paradigm on domains can therefore be reinterpreted as a reasoning paradigm over conceptual knowledge, with potential applications to symbolic data analysis.

To the best of our knowledge, the results in this paper constitute the first proposal for a default reasoning paradigm on conceptual knowledge which is compatible with mainstream research developments in nonmonotonic reasoning. We focus on laying foundations for this, but will not pursue questions of applicability to data analysis at this stage. This will be done elsewhere.

The plan of the paper is as follows. In Section 2 we recall main notions and results on the clausal logic of Rounds and Zhang, and its extension to a logic programming paradigm. In Section 3 we will add a notion of default negation, and in Section 4 we will see that it naturally extends answer set programming for extended disjunctive programs. Section 5 is devoted to the study of conceptual knowledge related to our paradigm. Related work is being discussed in Section 6, while we will conclude and discuss further work in Section 7.

Proofs have been omitted for lack of space; they can be found on the author's webpage.

*Acknowledgements.* This work was supported by a fellowship within the Postdoc-Programme of the German Academic Exchange Service (DAAD) and carried out while the author was visiting the Department of Electrical Engineering and Computer Science at Case Western Reserve University, Cleveland, Ohio. I am grateful for inspiring discussions with Rainer Osswald, Matthias Wendt, and Guo-Qiang Zhang, and for the feedback of some anonymous referees on an earlier version of this paper.

## 2   Clausal Logic and Logic Programming in Algebraic Domains

The study of domain theory from a logical perspective has a long tradition, and originates from [9], where a logical characterization (more precisely, a categorical equivalence) of bounded complete algebraic cpo's (with Scott continuous functions as morphisms) was given. Rounds and Zhang [1] have recently devised a similar characterization of Smyth powerdomains. They use a clausal logic for this purpose, and have also shown that it extends naturally to a disjunctive logic programming paradigm. We recall necessary notation and terminology in order to make this paper self-contained.

A *partially ordered set* is a pair $(D, \sqsubseteq)$, where $D$ is a nonempty set and $\sqsubseteq$ is a reflexive, antisymmetric, and transitive relation on $D$. A subset $X$ of a partially ordered set is *directed* if for all $x, y \in X$ there is $z \in X$ with $x, y \sqsubseteq z$. Note that the empty set is directed. An *ideal* is a directed and downward closed set. A *complete partial order*, *cpo* for short, is a partially ordered set $(D, \sqsubseteq)$ with a least element $\bot$, called the *bottom element* of $(D, \sqsubseteq)$, and such that every directed set in $D$ has a least upper bound, or supremum, $\bigsqcup D$. An element $c \in D$ is said to be *compact* or *finite* if whenever $c \sqsubseteq \bigsqcup L$ with $L$ directed, then there exists $e \in L$ with $c \sqsubseteq e$. The set of all compact elements of a cpo $D$ is written as $\mathsf{K}(D)$. An *algebraic cpo* is a cpo such that every $e \in D$ is the directed supremum of all compact elements below it. For $a, b \in D$ we write $a \mathbin{\not\uparrow} b$ if $a$ and $b$ are *inconsistent*, i.e. if there does not exist a common upper bound of $a$ and $b$.

A set $U \subseteq D$ is said to be *Scott open*, or just *open*, if it is upward closed and for any directed $L \subseteq D$ we have $\bigsqcup L \in U$ if and only if $U \cap L \neq \emptyset$. The *Scott topology* on $D$ is the topology whose open sets are all Scott open sets. An open set is *compact open* if it is compact in the Scott topology. A *coherent algebraic cpo* is an algebraic cpo such that the intersection of any two compact open sets is compact open. We will not make use of many topological notions in the sequel. So let us just note that coherency of an algebraic cpo implies that the set of all minimal upper bounds of a finite number of compact elements is finite, i.e. if $c_1, \ldots, c_n$ are compact elements, then the set $\mathsf{mub}\{c_1, \ldots, c_n\}$ of minimal upper bounds of these elements is finite. As usual, we set $\mathsf{mub}\,\emptyset = \{\bot\}$, where $\bot$ is the least element of $D$.

In the following, $(D, \sqsubseteq)$ will always be assumed to be a coherent algebraic cpo. We will also call these spaces *domains*. All of the above notions are standard and can be found e.g. in [2].

The following notions are taken from [1].

**Definition 1.** *Let $D$ be a coherent algebraic cpo with set $\mathsf{K}(D)$ of compact elements. A* clause *is a finite subset of $\mathsf{K}(D)$. We denote the set of all clauses over $D$ by $\mathcal{C}(D)$. If $X$ is a clause and $w \in D$, we write $w \models X$ if there exists $x \in X$ with $x \sqsubseteq w$, i.e. $X$ contains an element below $w$. A* theory *is a set of clauses, which may be empty. An element $w \in D$ is a* model *of a theory $T$, written $w \models T$, if $w \models X$ for all $X \in T$ or, equivalently, if every clause $X \in T$ contains an element below $w$. A clause $X$ is called a* logical consequence *of a theory $T$, written $T \models X$, if $w \models T$ implies $w \models X$. If $T = \{E\}$, then we write $E \models X$ for $\{E\} \models X$. Note that this holds if and only if for every $w \in E$ there is $x \in X$ with $x \sqsubseteq w$. For two theories $T$ and $S$, we say that $T \models S$ if $T \models X$ for all $X \in S$. In order to avoid confusion, we will throughout denote the empty clause by $\{\}$, and the empty theory by $\emptyset$. A theory $T$ is* closed *if $T \models X$ implies $X \in T$ for all clauses $X$. It is called* consistent *if $T \not\models \{\}$ or, equivalently, if there is $w$ with $w \models T$.*

The clausal logic introduced in Definition 1 will henceforth be called the *logic RZ* for convenience.

A main technical result from [1], where the notions from Definition 1 were introduced, shows that the set of all consistent closed theories over $D$, ordered by inclusion, is isomorphic to the collection of all non-empty Scott-compact saturated subsets of $D$, ordered by reverse inclusion — and the latter is isomorphic to the Smyth powerdomain of $D$. This result rests on the Hofmann-Mislove theorem [10]. It is also shown that a theory is logically closed if and only if it is an ideal,[1] and also that a clause is a logical consequence of a theory $T$ if and only if it is a logical consequence of a finite subset of $T$. The latter is a compactness theorem for clausal logic in algebraic domains.

*Example 1.* In [1], the following running example was given. Consider a countably infinite set of propositional variables, and the set $\mathbb{T} = \{\mathbf{f}, \mathbf{u}, \mathbf{t}\}$ of truth values ordered by $\mathbf{u} \leq \mathbf{f}$ and $\mathbf{u} \leq \mathbf{t}$. This induces a pointwise ordering on the space $\mathbb{T}^\mathcal{V}$ of all interpretations (or *partial truth assignments*). The partially ordered set $\mathbb{T}^\mathcal{V}$ is a coherent algebraic cpo[2] and has been studied e.g. in [11] in a domain-theoretic context, and in [12] in a logic programming context. Compact elements in $\mathbb{T}^\mathcal{V}$ are those interpretations which map all but a finite number of propositional variables to $\mathbf{u}$. We denote compact elements by strings such as $pq\overline{r}$, which indicates that $p$ and $q$ are mapped to $\mathbf{t}$ and $r$ is mapped to $\mathbf{f}$. Clauses in $\mathbb{T}^\mathcal{V}$ can be identified with formulae in disjunctive normal form, e.g. $\{pq\overline{r}, \overline{p}q, r\}$ translates to $(p \wedge q \wedge \neg r) \vee (\neg p \wedge q) \vee r$.

In [1], it was shown that the logic RZ is compact. A proof theory for it was also given. An alternative version was reported in [13,14].

The logic RZ provides a framework for reasoning with disjunctive information on a lattice which encodes background knowledge. Indeed it was shown [7] that it relates closely to formal concept analysis, which in turn has been applied successfully in data mining, and we will expand on this point later on in Section 5. Moreover, the system can be extended naturally to a disjunctive logic programming paradigm, as presented next, following [1].

**Definition 2.** *A (*disjunctive logic*) program over a domain $D$ is a set $P$ of rules of the form $Y \leftarrow X$, where $X, Y$ are clauses over $D$. An element $e \in D$ is said to be a* model *of $P$ if for every rule $Y \leftarrow X$ in $P$, if $e \models X$, then $e \models Y$. A clause $Y$ is a* logical consequence *of $P$ if every model of $P$ satisfies $Y$. We write* $\mathsf{cons}(P)$ *for the set of all clauses which are logical consequences of $P$. If $T$ is a theory, we write* $\mathsf{cons}(T)$ *for the set of all clauses which are logical consequences of $T$, i.e.* $\mathsf{cons}(T)$ *is the* logical closure *of $T$.*

Note that the notions of logical consequence differ for theories and programs. However, given a theory $T$, we have $\mathsf{cons}(T) = \mathsf{cons}(P_T)$, where $P_T = \{X \leftarrow \{\bot\} \mid X \in T\}$.

---

[1] An ideal with respect to the *Smyth preorder* $\sqsubseteq^\sharp$, where $X \sqsubseteq^\sharp Y$ if and only if for every $y \in Y$ there exists some $x \in X$ with $x \sqsubseteq y$.

[2] In fact it is also bounded complete.

The (*clause*) *propagation*[3] *rule*

$$\frac{X_1 \quad \ldots \quad X_n; \quad a_i \in X_i \quad (\text{all } i); \quad Y \leftarrow Z \in P; \quad \mathsf{mub}\{a_1,\ldots,a_n\} \models Z}{Y \cup \bigcup_{i=1}^{n}(X_i \setminus \{a_i\})},$$

denoted by $\mathsf{CP}(P)$, for given program $P$, was studied in [1]. Applying this rule, we say that $Y \cup \bigcup_{i=1}^{n}(X_i\setminus\{a_i\})$ is a $\mathsf{CP}(P)$-*consequence* of a theory $T$ if $X_1,\ldots,X_n \in T$. The following operator is based on the notion of $\mathsf{CP}(P)$-consequence and acts on logically closed theories. Let $T$ be a logically closed theory over $D$ and let $P$ be a program and define

$$\mathcal{T}_P(T) = \mathsf{cons}\left(\{Y \mid Y \text{ is a } \mathsf{CP}(P)\text{-consequence of } T\}\right).$$

In [1], it was shown that $\mathcal{T}_P$ is a Scott-continuous function on the space of all logically closed theories under set-inclusion, hence has a least fixed point $\mathsf{fix}(\mathcal{T}_P) = \bigsqcup\{\mathcal{T}_P \uparrow n\}$, where $\mathcal{T}_P \uparrow 0 = \mathsf{cons}(\{\{\bot\}\})$ and recursively $\mathcal{T}_P \uparrow (n+1) = \mathcal{T}_P(\mathcal{T}_P \uparrow n)$. It was also shown that $\mathsf{fix}(\mathcal{T}_P) = \mathsf{cons}(P)$.

## 3 Default Negation

We intend to add a notion of default negation to the logic programming framework presented above. The extension is close in spirit to mainstream developments concerning knowledge representation and reasoning with nonmonotonic logics.

**Definition 3.** *Let $D$ be a coherent algebraic domain. An* extended clause *is a pair $(C,N)$ of clauses over $D$, which we also write as "$C,\sim N$". An extended clause $(C,N)$ is called* trivially extended *if $N = \{\}$, and we may omit $N$ in this case. A (*trivially*) extended rule is of the form $Y \leftarrow X$, where $Y$ is a clause and $X$ is a (trivially) extended clause. An (*extended disjunctive*) program consists of a set of extended rules. If $Y \leftarrow C,\sim N$ is an extended rule, then we call $(C,N)$ the* body *of the rule and $Y$ the* head *of the rule.*

Informally, we read an extended rule $Y \leftarrow C,\sim N$ as follows: If $C$ holds, and $N$ does not, then $Y$ shall hold. This intuition gives rise to the following notions, akin to the answer set semantics [6], a point which we will discuss further in Section 4.

**Definition 4.** *Let $D$ be a coherent algebraic domain, let $P$ be an extended disjunctive program, and let $w \in D$. We define $P/w$ to be the (non-extended) program obtained by applying the following two transformations: (1) Replace each body $(C,N)$ of a rule by $C$ if $w \not\models N$. (2) Delete all rules with a body $(C,N)$ for which $w \models N$. An element $w \in D$ is an* answer model *of $P$ if it satisfies $w \models \mathsf{fix}\left(\mathcal{T}_{P/w}\right)$. An element $w \in D$ is a* min-answer model *of $P$ if it is minimal among all $v$ satisfying $v \models \mathsf{fix}\left(\mathcal{T}_{P/w}\right)$.*

---

[3] This rule was called the *hyperresolution rule determined by $P$* in [1].

Note that every min-answer model is an answer model. Recall also from [1] that the set of all models of a theory is compact saturated, hence is the upper closure of its minimal elements.

*Example 2.* Consider the (finite) domain $D$ depicted in Figure 1. This example is taken from [7] and encodes restaurant menues via formal concept analysis, a point which will be discussed in more detail later on in Section 5. We can now encode the wishes of a customer by programs, e.g. as follows.

$$\{d\} \leftarrow \{\bot\}$$
$$\{2,3,4\} \leftarrow \{\bot\}$$
$$\{rw\} \leftarrow \{\bot\}, \sim\{ww\}$$

Informally, the first rule states that the customer definitely wants a dessert. The second rule states that the customer wants one of the set meals 2, 3 or 4. The third rule states that the customer will choose red wine in all cases in which he does not have a good reason to choose white wine.

The element $4 \in D$ is a min-answer model for $P$, since $P/4$ consists of the clauses $\{d\} \leftarrow \{\bot\}$, $\{2,3,4\} \leftarrow \{\bot\}$, and $\{rw\} \leftarrow \{\bot\}$, and 4 is a minimal model of $\{\{d\}, \{2,3,4\}, \{rw\}\}$. Likewise, $3 \in D$ is a min-answer model since $P/3$ consists of the first two clauses from above and 3 is a minimal model of these. $7 \in D$ is an answer model of $P$, but not a min-answer model.

**Fig. 1.** Figure for Example 2. Abbreviations are: $sd$ salad, $st$ starter, $f$ fish, $m$ meat, $rw$ red wine, $ww$ white wine, $w$ water, $d$ dessert, $c$ coffee, $e$ expensive. Numbers 1 to 9 stand for set meals.

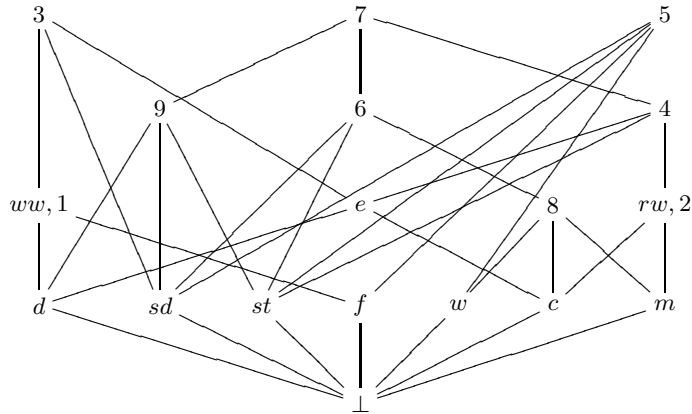

## 4  Answer Set Programming

*Answer set programming* is an artificial intelligent reasoning paradigm which was devised in order to capture some aspects of commonsense reasoning. More precisely, it is based on the observation that humans constantly tend to *jump to conlusions* in real-life situations, and on the idea that this imprecise reasoning mechanism (amongst other things) allows us to deal with the world effectively. Formally, *jumping to conclusions* can be studied by investigating supraclassical logics, see [15], where *supraclassicality* means, roughly speaking, that under such a logic more conclusions can be drawn from a set of axioms (or knowledge base) than could be drawn using classical (e.g. propositional or first-order) logic. Answer set programming, as well as the related default logic [3], is also non-monotonic, in the sense that a larger knowledge base does not necessarily yield a larger set of conclusions.

We next describe the notion of answer set for extended disjunctive logic programs, as proposed in [6]. It forms the heart of answer set programming systems like dlv [4], which have become a standard paradigm in artificial intelligence.

Let $\mathcal{V}$ denote a countably infinite set of propositional variables. A *rule* is an expression of the form

$$L_1, \ldots, L_n \leftarrow L_{n+1}, \ldots, L_m, \sim L_{m+1}, \ldots, \sim L_k,$$

where each of the $L_i$ is a literal, i.e. either a propositional variable or of the form $\neg p$ for some $p \in \mathcal{V}$. Given such a rule $r$, we set $\texttt{Head}(r) = \{L_1, \ldots, L_n\}$, $\texttt{Pos}(r) = \{L_{n+1}, \ldots, L_m\}$, and $\texttt{Neg}(r) = \{L_{m+1}, \ldots, L_k\}$.

In order to describe the answer set semantics, or stable model semantics, for extended disjunctive programs, we first consider programs without $\sim$. Thus, let $P$ denote an extended disunctive logic program in which $\texttt{Neg}(r)$ is empty for each rule $r \in P$. A subset $X$ of $\mathcal{V}^{\pm} = \mathcal{V} \cup \neg\mathcal{V}$ is said to be *closed by rules* in $P$ if, for every $r \in P$ such that $\texttt{Pos}(r) \subseteq X$, we have that $\texttt{Head}(r) \cap X \neq \emptyset$. The set $X \in 2^{\mathcal{V}^{\pm}}$ is called an *answer set* for $P$ if it is a minimal subset of $\mathcal{V}^{\pm}$ such that the following two conditions are satisfied.

1. If $X$ contains complementary literals, then $X = \mathcal{V}^{\pm}$.
2. $X$ is closed by rules in $P$.

We denote the set of answer sets of $P$ by $\alpha(P)$. Now suppose that $P$ is an extended disjunctive logic program that may contain $\sim$. For a set $X \in 2^{\mathcal{V}^{\pm}}$, consider the program $P/X$ defined as follows.

1. If $r \in P$ is such that $\texttt{Neg}(r) \cap X$ is not empty, then we remove $r$ i.e. $r \notin P/X$.
2. If $r \in P$ is such that $\texttt{Neg}(r) \cap X$ is empty, then the rule $r'$ belongs to $P/X$, where $r'$ is defined by $\texttt{Head}(r') = \texttt{Head}(r)$, $\texttt{Pos}(r') = \texttt{Pos}(r)$ and $\texttt{Neg}(r') = \emptyset$.

The program transformation $(P, X) \mapsto P/X$ is called the *Gelfond-Lifschitz transformation* of $P$ with respect to $X$.

It is clear that the program $P/X$ does not contain $\sim$ and therefore $\alpha\,(P/X)$ is defined. We say that $X$ is an *answer set* or *stable model* of $P$ if $X \in \alpha\,(P/X)$. So, answer sets of $P$ are fixed points of the operator $\mathsf{GL}_P$ introduced by Gelfond and Lifschitz in [6], where $\mathsf{GL}_P(X) = \alpha\,(P/X)$. We note that the operator $\mathsf{GL}_P$ is in general not monotonic, and call it the *Gelfond-Lifschitz operator* of $P$.

Now consider the coherent algebraic cpo $\mathbb{T}^{\mathcal{V}}$ from Example 1, and call an extended program over $\mathbb{T}^{\mathcal{V}}$ a *propositional program* if for each rule $Y \leftarrow (C, N)$ in $P$ we have that $Y$, $C$ and $N$ contain only atoms in $\mathbb{T}^{\mathcal{V}}$ or $\bot$, i.e. propositional variables or their negations (with respect to $\neg$) or $\bot$, $Y$ does not contain $\bot$, and $C$ is a singleton clause.

Now if $P$ is a propositional program, then let $P'$ be the extended disjunctive logic program obtained from $P$ by transforming each rule

$$\{p_1, \ldots, p_n\} \leftarrow \{q_1 \ldots q_m\}, \sim\{r_1, \ldots, r_k\}$$

from $P$ into the rule

$$p_1, \ldots, p_n \leftarrow q_1, \ldots, q_m, \sim r_1, \ldots, \sim r_k,$$

in $P'$, where $p_i, q_j, r_l$ are atoms in $\mathbb{T}^{\mathcal{V}}$, i.e. literals over $\mathcal{V}$. If $q_1 \ldots q_m = \bot$, then it is omitted. If $r_i = \bot$ for some $i$, then the rule will never play a role, so we can assume without loss of generality that this does not occur. This transformation can obviously be reversed. We say that $P$ and $P'$ are *associated with each other*.

**Theorem 1.** *Let $P$ be a propositional program and $P'$ be its associated extended disjunctive logic program. Then $w \in \mathbb{T}^{\mathcal{V}}$ is a min-answer model of $P$ if and only if $w' = \{p \in \mathcal{V}^{\pm} \mid w \models \{p\}\}$, i.e. the set of all atoms for which $w$ is a model, is an answer set for $P'$. Conversely, if $X \subseteq \mathcal{V}^{\pm}$ is an answer set for $P'$ which does not contain complementary literals, then $x = \bigsqcup X \in \mathbb{T}^{\mathcal{V}}$ exists and is a min-answer model of $P$. If $\mathcal{V}^{\pm}$ is an answer set for $P'$ (and hence the only answer set of $P'$), then $P$ does not have any min-answer models.*

Theorem 1 shows that reasoning (or programming) with min-answer models encompasses answer set programming with extended disjunctive logic programs. More precisely, we obtain the classical answer set programming paradigm by restricting our attention to the domain $\mathbb{T}^{\mathcal{V}}$. What do we gain through this more general framework? One the one hand, we improve in conceptual clarity: Our results open up the possibility of a domain-theoretical (and domain-logical) treatment of answer set programming in the basic paradigm, and possibly also for some extensions recently being studied. On the other hand, we gain flexibility due to the possible choice of underlying domain, which we like to think of as background knowledge on which we program or which we query. The choice of $\mathbb{T}^{\mathcal{V}}$ corresponds to the language of propositional logic, and all order structures satisfying the requirements of being coherent algebraic cpos are suitable. These requirements are rather weak from a computational perspective, because among the computationally relevant order structures studied in domain theory, coherent algebraic cpos form a rather general class. In particular, they encompass all finite partial orders, and all complete algebraic lattices.

In Section 5 we will actually propose a very general way — using formal concept analysis — of obtaining suitable order structures.

The following theorem is an immediate corollary from Theorem 1.

**Theorem 2.** *Let $P$ be a propositional program not containing $\sim$ and $P'$ be its associated extended disjunctive program. If $w \in \mathbb{T}^{\mathcal{V}}$ is a minimal model of $P$ then $w' = \{p \in \mathcal{V}^{\pm} \mid w \models \{p\}\}$ is minimally closed by rules in $P'$. Conversely, if $X \subseteq \mathcal{V}^{\pm}$ is minimally closed by rules in $P'$ and does not contain complementary literals, then $x = \bigsqcup X \in \mathbb{T}^{\mathcal{V}}$ exists and is a minimal model of $P$. If $\mathcal{V}^{\pm}$ is minimally closed by rules in $P'$ (and thus is the only answer set of $P'$), then $P$ does not have any models.*

In particular, Theorem 1 shows that the minimal model semantics for definite logic programs [16] can be recovered using the original approach from [1] without default negation. Likewise, the same holds for the stable model semantics for normal logic programs [17], which are non-disjunctive ones without negation $\neg$.

## 5  Formal Concept Analysis

Formal concept analysis is a powerful lattice-based approach to symbolic data analysis. It was devised in the 1980s [18] and was originally inspired by ideas from philosophy, more precisely by *Port Royal Logic*, which describes a *concept* as consisting of a set of objects (the *extent* of the concept) and a set of attributes (the *intent* of the concept) such that these objects share exactly all these attributes and vice-versa. In the meantime, an active community is driving the field, covering mathematical foundations, logical aspects, and applications in data mining, ontology engineering, artificial intelligence, and elsewhere.

The formation of concepts can be viewed as logical closure in the sense that a set of attributes $B$ implies an attribute $m$ (which may or may not be contained in $B$), if all objects which fall under all attributes in $B$ also share the attribute $m$. This will be made more precise below. We thus obtain a notion of logical consequence on attribute sets, respectively a natural implicative theory, which corresponds to so-called *association rules* in data mining. This implicative theory is intimately related to the logic RZ, a point which we mentioned earlier and will study formally in the following. The strong correspondence between the logic RZ and the formation of formal concepts from formal contexts has already been reported in [7] for the case of finite contexts. We will now supplement these results by a theorem which treats the case of infinite contexts.

We first introduce the notions of formal context and concept as used in formal concept analysis. We follow the standard reference [8].

A (*formal*) *context* is a triple $(G, M, I)$ consisting of two sets $G$ and $M$ and a relation $I \subseteq G \times M$. Without loss of generality, we assume that $G \cap M = \emptyset$. The elements of $G$ are called the *objects* and the elements of $M$ are called the *attributes* of the context. For $g \in G$ and $m \in M$ we write $gIm$ for $(g, m) \in I$, and say that *g has the attribute m*.

For a set $A \subseteq G$ of objects we set $A' = \{m \in M \mid gIm \text{ for all } g \in A\}$, and for a set $B \subseteq M$ of attributes we set $B' = \{g \in G \mid gIm \text{ for all } m \in B\}$. A (*formal*) *concept* of $(G, M, I)$ is a pair $(A, B)$ with $A \subseteq G$ and $B \subseteq M$, such that $A' = B$ and $B' = A$. We call $A$ the *extent* and $B$ the *intent* of the concept $(A, B)$. For singleton sets, e.g. $B = \{b\}$, we simplify notation by writing $b'$ instead of $\{b\}'$.

The set $\mathcal{B}(G, M, I)$ of all concepts of a given context $(G, M, I)$ is a complete lattice with respect to the order defined by $(A_1, B_1) \leq (A_2, B_2)$ if and only if $A_1 \subseteq A_2$, which is equivalent to the condition $B_2 \subseteq B_1$. $\mathcal{B}(G, M, I)$ is called the *concept lattice* of the context $(G, M, I)$.

*Remark 1.* For every set $B \subseteq M$ of attributes we have that $B' = B'''$, so that $(B', B'')$ is a concept. Hence, the concept lattice of a context $(G, M, I)$ can be identified with the set $\{B'' \mid B \subseteq M\}$, ordered by reverse subset inclusion.

Furthermore, if $m \in M$ is an attribute, then we call $(m', m'') = (\{m\}', \{m\}'')$ an *attribute concept*. Dually, if $g \in G$ is an object, then we call $(g'', g') = (\{g\}'', \{g\}')$ an *object concept*. The subposet $L$ of $\mathcal{B}(G, M, I)$ consisting of all attribute and object concepts is called the *Galois subhierarchy* or *AOC* associated with $(G, M, I)$. By abuse of notation, we denote members of $L$ by elements from $G \cup M$. This is justified by the obvious possibility to identify the set $L$ with $(G \cup M)/\sim$, where $\sim$ is the equivalence relation identifying each two elements in $G \cup M$ whose associated concepts coincide. We denote the induced order on $L$ by $\leq$.

**Theorem 3.** *Let $(G, M, I)$ be a formal context, $\mathcal{B}(G, M, I)$ be the corresponding formal concept lattice, and $(L, \leq)$ be the Galois subhierarchy associated with $(G, M, I)$. Let $(D, \sqsubseteq)$ be a coherent algebraic cpo and $\iota : L \to D$ be an order-reversing injective function which covers all of $\mathsf{K}(D)$, i.e. for each $c \in \mathsf{K}(D)$ there exists some $a \in L$ with $\iota(a) = c$. Furthermore, let $A = \{m_1, \ldots, m_n\} \subseteq M$ such that $\iota(m_i) \in \mathsf{K}(D)$ for all $i$. Then*

$$A'' = \{m \mid \{\{\iota(m_1)\}, \ldots, \{\iota(m_n)\}\} \models \{\iota(m)\}\}.$$

We remark that Theorem 3 applies to all finite contexts since the Galois subhierarchy of a finite context is always (and trivially) a coherent algebraic cpo where all elements are compact; a bottom element may have to be added, though. This finite case is also a corollary from [7, Theorem 3], taking Example 1 and Proposition 1 from [7] into account.

The following example is taken from [7]; it complements Example 2.

*Example 3.* Consider the formal context given in Table 1. It shall represent, in simplified form, a selection of set dinners from a restaurant menu. The Galois subhierarchy of its formal concept lattice is depicted in Figure 1. Concepts in this setting correspond to types of dinners, e.g. one may want to identify the concept with extent $\{4, 6, 7\}$ and intent $\{st, m, c\}$, using the abbreviations from Figure 1, to be the *heavy* meals, while the *expensive* ones are represented by the attribute concept of $e$, and turn out to always include coffee. Using the logic

Table 1. Formal context for Example 3.

|   | salad | starter | fish | meat | red wine | white wine | water | dessert | coffee | expensive |
|---|---|---|---|---|---|---|---|---|---|---|
| 1 |   |   | × |   |   | × |   | × |   |   |
| 2 |   |   |   | × | × |   |   |   | × |   |
| 3 | × |   | × |   |   | × |   | × | × | × |
| 4 |   | × |   | × | × |   |   | × | × | × |
| 5 | × | × | × |   |   |   | × |   |   |   |
| 6 | × | × |   | × |   |   | × |   | × |   |
| 7 | × | × |   | × | × |   | × | × | × | × |
| 8 |   |   |   | × |   |   | × |   | × |   |
| 9 | × | × |   |   |   |   |   |   | × |   |

RZ, we can for example conclude that a customer who wants salad and fish will choose one of the meals 3 or 5, since these elements of the poset are exactly those which are both objects and models of the theory $\{\{sd\},\{f\}\}$. Also, he will always get a starter or a dessert, formally $\{\{sd\},\{f\}\} \models \{st,d\}$. To give a slightly more sophisticated example, suppose that a customer wants salad or a starter, additionally fish or a dessert, and drinks water. From this we can conclude that in any case he will get both a salad and a starter. Formally, we obtain $\{\{sd,st\},\{f,d\},\{w\}\} \models \{sd\}$ and $\{\{sd,st\},\{f,d\},\{w\}\} \models \{st\}$. A little bit of reflection on the context makes it clear that these inferences are indeed natural ones.

Let us stop for a moment and dwell on the significance of Theorem 3. We note first of all that the hypothesis is not very strong from a domain-theoretic perspective: we encompass all concept lattices for which some corresponding Galois subhierarchy forms at least an abstract basis for a coherent algebraic cpo. One could argue that such or similar conditions have to be satisfied in any case if one intends to perform computation on an infinite order structure. The conclusion of the theorem then says that concept closure (or in other words, the underlying implicative theory) basically coincides with consequence in RZ, restricted to finite sets of singleton clauses, which can be interpreted as conjunctions of elements or items from $G \cup M$. The logic RZ then lifts concept closure to become part of disjunctive reasoning, in a natural and intuitively appealing way. From this perspective we can say that the logic RZ *is* the implicative theory obtained from concept closure, naturally extended with a notion of disjunction.

What we gain from this perspective is not only a tight relationship between formal concept analysis and domain theory, but also a non-monotonic reasoning paradigm on conceptual knowledge, by utilizing our results in Section 4. Formal contexts can now be interpreted as providing *background knowledge* in elementary form, which can be queried, or programmed on, by using disjunctive logic programs with default negation, as described in Section 4. From this, we obtain a clear distinction between the (monotonic!) background knowledge or underlying database, and the program written on top of it, allowing for a clear separation of the nonmonotonic aspects which are diffcult to deal with efficiently and effectively.

## 6   Related Work

Logical aspects of formal concept analysis have certainly received ample attention in the literature, see e.g. [19,20]. In particular, the *contextual attribute logic* due to Ganter and Wille [19] is closely related to our results in Section 5, and for the finite case this was spelled out in [7].

The study of relationships between formal concept analysis and domain theory has only recently received attention. Zhang and Shen [21,22] approach the issue from the perspective of Chu spaces and Scott information systems. A category-theoretical setting was developed from these investigations in [23]. The work just mentioned has a different focus than our result in Section 5 and [7], but develops along similar basic intuitions and is mainly compatible with ours. Its flavour is more category-theoretical and targets categorial constructions which may be used for ontology engineering.

Osswald and Petersen [24,25] study an approach to encoding knowledge in order structures which is inspired from linguistics. They obtain a framework which is more flexible than formal concept analysis, and appears to be compatible with our results in Section 5 and [7]. They also propose a default reasoning paradigm, but it remains to be worked out how it relates to ours.

Relationships between domain theory and nonmonotonic reasoning have hardly been studied in the literature, except from series of papers by Rounds and Zhang, e.g. [1,26,27], and Hitzler and Seda, e.g. [28,29,30]. This is remarkable since domain theory has become a respected paradigm in the theory of computing with widespread applications. We believe that this relationship deserves much more attention in order to understand the theoretical underpinnings of nonmonotonic reasoning and other artificial intelligence paradigms.

Default reasoning on concept hierarchies has also been studied before, for example in the form of default reasoning in semantic networks, e.g. [31], and as nonmonotonic reasoning with ontologies, e.g. [32,33]. Since ontology creation is a currently evolving area of application for formal concept analysis, we expect that our paradigm will also be useful for similar purposes. Another related paradigm is logic programming with inheritance [34], where the underlying order structures are is-a hierarchies, which do not have a similarly rich logical structure as the logic RZ or Galois subhierarchies of formal concept lattices.

## 7   Conclusions and Further Work

The work presented in this paper touches domain theory, nonmonotonic reasoning, and symbolic data analysis. The contribution should mainly be considered as an inspiration for further investigations which grow naturally out of our observations. There are several starting points for such work, and some of them bear potential for full research projects which are interesting in their own right.

Concerning the relations worked out between the logic RZ and nonmonotonic reasoning, we have described a general reasoning framework which encompasses answer set programming with extended disjunctive programs as a special case, namely with the domain restricted to $\mathbb{T}^\mathcal{V}$. This opens up new ways for domain-theoretic analysis for nonmonotonic reasoning in this paradigm, with the hope that e.g. decidability aspects could be tackled — an issue which has so far received only little attention in the nonmonotonic reasoning community. On the other hand, by substituting $\mathbb{T}^\mathcal{V}$ by other domains, it should be possible to lift answer set programming out of the restricted syntax provided by the fragment of first-order logic usually considered.

Concerning the relations between the logic RZ and formal concept analysis displayed in Section 5, we can understand the logic RZ as a means of reasoning with conceptual knowledge, related to the approach presented in [19], as already mentioned in [7]. Indeed, the choice of $\mathbb{T}^\mathcal{V}$ as underlying domain relates to answer set programming, while the choice of other domains can be motivated by formal concept analysis. Of particular interest are also the infinitary aspects of this, and the potential of the domain-theoretic approach to deal with questions of computability and query-answering even on infinite contexts. From this perspective, it should be investigated under which conditions a context satisfies the hypotheses of Theorem 3. It would also be important to relate this result to those of [21], where domain theory and formal concept analysis are being related by means of Chu space theory, and [24,25], where a general approach encompassing formal concept analysis is described for obtaining order structures carrying hierarchical knowledge.

Finally, we would like to emphasize that the results presented here lead to a nonmonotonic reasoning paradigm on conceptual knowledge. More precisely, starting from a given (and possibly infinite) context, we have provided means for doing nonmonotonic reasoning on the Galois subhierarchy of the context. Since the logic RZ captures the notion of concept closure, we obtain a reasoning paradigm dealing with conceptual knowledge in a way very natural to formal concept analysis. On the other hand, the nonmonotonic reasoning paradigm thus put in place is very close in spirit to mainstream developments in answer set programming, and can thus benefit from the experience gained within this field of research.

We believe that the resulting nonmonotonic reasoning paradigm with conceptual knowledge bears potential for applications. One could envisage background knowledge in the form of formal contexts, and sophisticated queries or planning tasks expressed by programs. We are not aware of any other work which proposes a default reasoning paradigm on conceptual knowledge compatible with mainstream research developments in nonmonotonic reasoning.

## Appendix: Proofs

**Theorem 1** Let $P$ be a propositional program and $P'$ be its associated extended disjunctive logic program. Then $w \in \mathbb{T}^\mathcal{V}$ is a min-answer model of $P$ if and only if $w' = \{p \in \mathcal{V}^\pm \mid w \models \{p\}\}$, i.e. the set of all atoms for which $w$ is a model, is an answer set for $P'$. Conversely, if $X \subseteq \mathcal{V}^\pm$ is an answer set for $P'$ which does not contain complementary literals, then $x = \bigsqcup X \in \mathbb{T}^\mathcal{V}$ exists and is a min-answer model of $P$. If $\mathcal{V}^\pm$ is an answer set for $P'$ (and hence the only answer set of $P'$), then $P$ does not have any min-answer models.

*Proof.* Note that $(P/w)' = (P'/w')$, so we first restrict our attention to programs without $\sim$. Let $Q, Q'$ be such programs associated with each other. We show first that for every min-answer model $v$ for $Q$ we have that

$$v' = \{p \in \mathcal{V}^\pm \mid v \models \{p\}\}$$

is closed by rules in $Q'$. Then we show that for every answer set $X$ of $Q'$ we have that $x = \bigsqcup X$ is an answer model of $Q$. Then we proceed with showing that $v'$ is an answer set if $v$ is a min-answer model, and finally that $x = \bigsqcup X$ is a min-answer model for $Q$ whenever $X$ is an answer set for $Q'$.

So let $v \in \mathbb{T}^\mathcal{V}$ be a min-answer model for $Q$. Then $v$ is minimal among all $z$ with $z \models \mathsf{fix}(\mathcal{T}_{Q/v}) = \mathsf{fix}(\mathcal{T}_Q) = \mathsf{cons}(Q)$, i.e. $v$ is a minimal model of $Q$. Note that $v = \bigsqcup v' = \bigsqcup \{p \in \mathcal{V}^\pm \mid v \models \{p\}\}$, so $v'$ cannot contain complementary literals, since two complementary literals do not have a common upper bound in $\mathbb{T}^\mathcal{V}$. We show next that $v'$ is closed by rules in $Q'$. So let $p_1, \ldots, p_n \leftarrow q_1, \ldots, q_m$ be a rule in $Q'$ such that $q_1, \ldots, q_m \in v'$. But then $v \models \bigsqcup\{q_1, \ldots, q_m\} = q_1 \ldots q_m$ and $\{p_1, \ldots, p_n\} \leftarrow \{q_1 \ldots q_m\}$ is a clause in $Q$. Since $v$ is a model of $Q$ we obtain $v \models \{p_1, \ldots, p_n\}$ and hence $v \models \{p_i\}$ for some $i$, which implies $p_i \in v'$ as desired.

Now let $X \subseteq \mathcal{V}^\pm$ be an answer set for $Q'$ which does not contain complementary literals. Then $x = \bigsqcup X$ exists, and we show that it is an answer model of $Q$. So let $\{p_1, \ldots, p_n\} \leftarrow \{q_1 \ldots q_m\}$ be a clause in $Q$ with $x \models \{q_1 \ldots q_m\}$. Then $\{q_1, \ldots, q_m\} \subseteq X$, hence $p_i \in X$ for some $i$, and consequently $x \models \{p_i\}$ as desired.

For a min-answer model $v$ for $Q$ we know already that $v'$ does not contain complementary literals and is closed by rules in $Q'$. Now let $Y \subseteq v'$ be an answer set for $Q'$. Then $\bigsqcup Y$ exists and $\bigsqcup Y \sqsubseteq \bigsqcup v' = v$, hence $\bigsqcup Y = v$ by minimality of $v$, and consequently $Y = v'$, so $v'$ is an answer set for $Q'$.

For an answer set $X$ for $Q'$ we know already that $x = \bigsqcup X$ is an answer model of $Q$. Now let $y \sqsubseteq x$ be a min-answer model for $Q$. Then $y' = \{p \in \mathcal{V}^\pm \mid y \models \{p\}\} \subseteq X$ is an answer set for $Q'$, hence $y' = X$ by minimality of $X$, and consequently $y = x$, so $x$ is a min-answer model for $Q$.

This closes the proof for programs without $\sim$. Now let $P$ be a propositional program including $\sim$.

Let $w$ be a min-answer model for $P$, hence $w$ is minimal among all $v$ with $v \models \mathsf{fix}(\mathcal{T}_{P/w})$, and in particular $w$ is a min-answer model for $P/w$. So $w' = \{p \in \mathcal{V}^\pm \mid w \models \{p\}\}$ is an answer set for $(P/w)' = P'/w'$ as desired.

Let $X$ be an answer set for $P'$ which does not contain complementary literals. Then $X$ in particular is an answer set for $P'/X = (P/\bigsqcup X)'$, hence $x = \bigsqcup X$ is a min-answer model for $P/x$, and consequently also a min-answer set model $P$ as desired.

If $X$ is an answer set for $P'$ containing complementary literals then $X = \mathcal{V}^\pm$, and $X$ is the only answer set for $P'$. Now if $P$ had a min-answer model $w$, then $w' = \{p \in \mathcal{V}^\pm \mid w \models \{p\}\} \subseteq X$ were an answer set for $P$. But then $w' = X$ which is impossible since this implies $w = \bigsqcup X$, but the supremum does not exist.

**Theorem 3** Let $(G, M, I)$ be a formal context, $\mathcal{B}(G, M, I)$ be the corresponding formal concept lattice, and $(L, \leq)$ be the Galois subhierarchy associated with $(G, M, I)$. Let $(D, \sqsubseteq)$ be a coherent algebraic cpo and $\iota : L \to D$ be an order-reversing injective function which covers all of $\mathsf{K}(D)$, i.e. for each $c \in \mathsf{K}(D)$ there exists some $a \in L$ with $\iota(a) = c$. Furthermore, let $A = \{m_1, \ldots, m_n\} \subseteq M$ such that $\iota(m_i) \in \mathsf{K}(D)$ for all $i$. Then

$$A'' = \{m \mid \{\{\iota(m_1)\}, \ldots, \{\iota(m_n)\}\} \models \{\iota(m)\}\}.$$

*Proof.* Let $m$ be such that $\{\{\iota(m_1)\}, \ldots, \{\iota(m_n)\}\} \models \{\iota(m)\}$. We have to show that $m \in A''$. So let $g \in G$ be such that $gIm_i$ for all $i$, which implies $g \leq m_i$ for all $i$. So $\iota(g) \sqsupseteq \iota(m_i)$ for all $i$, i.e. $\iota(g) \models \{\{\iota(m_1)\}, \ldots, \{\iota(m_n)\}\}$, and consequently $\iota(g) \models \{\iota(m)\}$. It follows that $\iota(g) \sqsupseteq \iota(m)$ which implies $g \leq m$, hence $gIm$. Since $g$ was chosen arbitrarily, we conclude $m \in A''$.

Conversely, let $m \in A''$ and let $w$ be chosen arbitrarily with $w \in \mathsf{mub}\,\iota(A)$. Then it remains to show that $\iota(m) \sqsubseteq w$ since this implies $\iota(m) \sqsubseteq x$ for all $x$ with $x \models \{\{\iota(m_1)\}, \ldots, \{\iota(m_n)\}\}$ by arbitrary choice of $w$, and hence

$$\{\{\iota(m_1)\}, \ldots, \{\iota(m_n)\}\} \models \{\iota(m)\}$$

as desired.

In order to show $\iota(m) \sqsubseteq w$ first note that $w \in \mathsf{K}(D)$ by coherency and our assumption on $A$. We consider the two cases (i) $\iota^{-1}(w) \in G$ and (ii) $\iota^{-1}(w) \in M$.

(i) If $\iota^{-1}(w) = g \in G$, then $g \leq m_i$ for all $i$, hence $gIm_i$ for all $i$ and consequently $gIm$. We obtain $g \leq m$, hence $w = \iota(g) \sqsupseteq \iota(m)$ as desired.

(ii) If $\iota^{-1}(w) = a \in M$, then $a \leq m_i$ for all $i$, hence $a' \subseteq m'_i$ for all $i$. So for all $g$ with $gIa$ we have $gIm_i$ for all $i$, and by $m \in A''$ we obtain $gIm$. We have just shown that $g \in a'$ implies $g \in m'$, so $a' \subseteq m'$, which is equivalent to $a \leq m$. Hence $w = \iota(a) \sqsupseteq \iota(m)$ as desired.